\documentclass{article}
\usepackage{ijcai16}
\usepackage{times}
\usepackage{url}
\usepackage{arydshln}
\usepackage{example}
\usepackage{graphicx}
\usepackage{amsmath}
\usepackage{amssymb}
\usepackage{multirow}
\DeclareMathOperator*{\argmax}{argmax}
\DeclareMathOperator*{\argmin}{argmin}

\title{Solving Verbal Questions in IQ Test by Knowledge-Powered Word Embedding}
\author{
Huazheng Wang\\
The University of Virginia\\
huazhengwang@gmail.com
\And
Fei Tian\\
University of Science and Technology of China\\
tianfei@mail.ustc.edu.cn
\AND
Bin Gao\\
Microsoft Research\\
bingao@microsoft.com
\And
Jiang Bian\\
Yidian Inc\\
jiang.bian.prc@gmail.com
\And
Tie-Yan Liu\\
Microsoft Research\\
tyliu@microsoft.com
}

\begin{document}

\maketitle

\begin{abstract}
  Verbal comprehension questions appear very frequently in Intelligence Quotient (IQ) tests, which measure human's verbal ability including the understanding of the words with multiple senses, the synonyms and antonyms, and the analogies among words. In this work, we explore whether such tests can be solved automatically by the deep learning technologies for text data. We found that the task was quite challenging, and simply applying existing technologies like word embedding could not achieve a good performance, due to the multiple senses of words and the complex relations among words. To tackle these challenges, we propose a novel framework to automatically solve the verbal IQ questions by leveraging improved word embedding by jointly considering the multi-sense nature of words and the relational information among words. Experimental results have shown that the proposed framework can not only outperform existing methods for solving verbal comprehension questions but also exceed the average performance of the Amazon Mechanical Turk workers involved in the study.
\end{abstract}

\section{Introduction}\label{sec-introduction}

The Intelligence Quotient (IQ) test~\cite{William1914} is a test of intelligence designed to formally study the success of an individual in adapting to a specific situation under certain conditions. Common IQ tests measure various types of abilities such as verbal, mathematical, logical, and reasoning skills. These tests have been widely used in the study of psychology, education, and career development. In the community of artificial intelligence, agents have been invented to fulfill many interesting and challenging tasks like face recognition, speech recognition, handwriting recognition, and question answering. However, as far as we know, there are very limited studies of developing an agent to solve IQ tests, which in some sense is more challenging, since even common human beings could not always succeed in the tests. Considering that IQ test scores have been widely considered as a measure of \emph{intelligence}, we think it is worth making further investigations whether we can develop an agent that can solve IQ tests.

The commonly used IQ tests contain several types of questions like verbal, mathematical, logical, and picture questions, among which a large proportion (near 40\%) are verbal questions~\cite{Carter2005}. The recent progress on deep learning for natural language processing (NLP), such as word embedding technologies, has advanced the ability of machines (or AI agents) to understand the meaning of words and the relations among words. This inspires us to solve the verbal questions in IQ tests by leveraging the word embedding technologies. However, our attempts show that a straightforward application of word embedding could not result in satisfactory performances. This is actually understandable. Standard word embedding technologies learn one embedding vector for each word based on the co-occurrence information in a text corpus. However, verbal comprehension questions in IQ tests usually consider the multiple senses of a word (and often focus on the rare senses), and the complex relations among (polysemous) words. This has clearly exceeded the capability of standard word embedding technologies.

To tackle the aforementioned challenges, we propose a novel framework that consists of three components.

First, we build a classifier to recognize the specific type (e.g., analogy, classification, synonym, and antonym) of verbal questions. For different types of questions, different kinds of relationships need to be considered and the solvers could have different forms. Therefore, with an effective question type classifier, we may solve the questions in a divide-and-conquer manner.

Second, we obtain distributed representations of words and relations by leveraging a novel word embedding method that considers the multi-sense nature of words and the relational knowledge among words (or their senses) contained in dictionaries. In particular, for each polysemous word, we retrieve its number of senses from a dictionary, and conduct clustering on all its context windows in the corpus. Then we attach the example sentences for every sense in the dictionary to the clusters, such that we can tag the polysemous word in each context window with a specific word sense. On top of this, instead of learning one embedding vector for each \emph{word}, we learn one vector for each \emph{pair of word-sense}. Furthermore, in addition to learning the embedding vectors for words, we also learn the embedding vectors for relations (e.g., synonym and antonym) at the same time, by incorporating relational knowledge into the objective function of the word embedding learning algorithm. That is, the learning of word-sense representations and relation representations interacts with each other, such that the relational knowledge obtained from dictionaries is effectively incorporated.

Third, for each type of questions, we propose a specific solver based on the obtained distributed word-sense representations and relation representations. For example, for analogy questions, we find the answer by minimizing the distance between word-sense pairs in the question and the word-sense pairs in the candidate answers.

We have conducted experiments using a combined IQ test set to test the performance of our proposed framework. The experimental results show that our method can outperform several baseline methods for verbal comprehension questions in IQ tests. We further deliver the questions in the test set to human beings through Amazon Mechanical Turk\footnote{\url{http://www.mturk.com/}}. The average performance of the human beings is even a little lower than that of our proposed method.

\section{Related Work}\label{sec-related-work}

\subsection{Verbal Questions in IQ Test}

In common IQ tests, a large proportion of questions are verbal comprehension questions, which play an important role in deciding the final IQ scores. For example, in Wechsler Adult Intelligence Scale~\cite{wechsler2008wechsler}, which is among the most famous IQ test systems, the full-scale IQ is calculated from two IQ scores: Verbal IQ and Performance IQ, and around 40\% questions in a typical test are verbal comprehension questions. Verbal questions can test not only the verbal ability (e.g., understanding polysemy of a word), but also the reasoning ability and induction ability of an individual. According to previous studies~\cite{Carter2005}, verbal questions mainly have the types elaborated in Table \ref{table:type}, in which the correct answers are highlighted in bold font.

\begin{table*}
  \centering\tiny
  \vspace{-5pt}
  \caption{Types of verbal questions.}\label{table:type}
  \begin{tabular}{c|l}
  \hline
  \hline
  Type & Example \\
  \hline
  Analogy-I & Isotherm is to temperature as isobar is to? (i) atmosphere, (ii) wind, (iii) \textbf{pressure}, (iv) latitude, (v) current. \\
  \hline
  Analogy-II & Identify two words (one from each set of brackets) that form a connection (analogy) when paired with the words in capitals: CHAPTER (\textbf{book}, verse, read), ACT (stage, audience, \textbf{play}). \\
  \hline
  Classification & Which is the odd one out? (i) calm, (ii) \textbf{quiet}, (iii) relaxed, (iv) serene, (v) unruffled. \\
  \hline
  Synonym & Which word is closest to IRRATIONAL? (i)intransigent, (ii) irredeemable, (iii) unsafe, (iv) lost, (v) \textbf{nonsensical}. \\
  \hline
  Antonym & Which word is most opposite to MUSICAL? (i) \textbf{discordant}, (ii) loud, (iii) lyrical, (iv) verbal, (v) euphonious. \\
  \hline
  \end{tabular}\vspace{-15pt}
\end{table*}

Analogy-I questions usually take the form ``$A$ is to $B$ as $C$ is to ?''. One needs to choose a word $D$ from a given list of candidate words to form an analogical relation between pair ($A$, $B$) and pair ($C$, $D$). Such questions test the ability of identifying an implicit relation from word pair ($A$, $B$) and apply it to compose word pair ($C$, $D$). Note that the Analogy-I questions are also used as a major evaluation task in the \emph{word2vec} models~\cite{mikolov2013distributed}. Analogy-II questions require two words to be identified from two given lists in order to form an analogical relation like ``$A$ is to ? as $C$ is to ?''. Such questions are a bit more difficult than the Analogy-I questions since the analogical relation cannot be observed directly from the questions, but need to be searched in the word pair combinations from the candidate answers. Classification questions require one to identify the word that is different (or dissimilar) from others in a given word list. Such questions are also known as \emph{odd-one-out}, which have been studied in~\cite{Pinter2012}. Classification questions test the ability of summarizing the majority sense of the words and identifying the outlier. Synonym questions require one to pick one word out of a list of words such that it has the closest meaning to a given word. Synonym questions test the ability of identifying all senses of the candidate words and selecting the correct sense that can form a synonymous relation to the given word. Antonym questions require one to pick one word out of a list of words such that it has the opposite meaning to a given word. Antonym questions test the ability of identifying all senses of the candidate words and selecting the correct sense that can form an antonymous relation to the given word.

Although there are some efforts to solve mathematical, logical, and picture questions in IQ test~\cite{Sanghi2003,Strannegard2012-1,Kushmany2014,seo2014aaai,hosseini2014emnlp,weston2015towards}, there has been very few efforts to develop automatic methods to solve verbal questions.

\subsection{Deep Learning for Text Mining}

Building distributed word representations~\cite{bengio2003neural}, a.k.a. word embeddings, has attracted increasing attention in the area of machine learning. Different with conventional \emph{one-hot} representations of words or distributional word representations based on co-occurrence matrix between words such as LSA~\cite{dumais1988LSA} and LDA~\cite{blei2003latent}, distributed word representations are usually low-dimensional dense vectors trained with neural networks by maximizing the likelihood of a text corpus. Recently, to show its effectiveness in a variety of text mining tasks, a series of works applied deep learning techniques to learn high-quality word representations~\cite{collobert2008unified,mikolov2013distributed,pennington2014glove}.

Nevertheless, since the above works learn word representations mainly based on the word co-occurrence information, it is quite difficult to obtain high quality embeddings for those words with very little context information; on the other hand, large amount of noisy or biased context could give rise to ineffective word embeddings either. Therefore, it is necessary to introduce extra knowledge into the learning process to regularize the quality of word embedding. Some efforts have paid attention to learn word embedding in order to address knowledge base completion and enhancement~\cite{bordes2011learning,socher2013reasoning,weston2013connecting}, and some other efforts have tried to leverage knowledge to enhance word representations~\cite{luong2013better,WestonBYU13,FriedD14,Celikyilmaz15}. Moreover, all the above models assume that one word has only one embedding no matter whether the word is polysemous or monosemous, which might bring some confusion for the polysemous words. To solve the problem, there are several efforts like~\cite{EricHuang2012,Tian2014,neelakantan-EtAl2014Nonparametric}. However, these models do not leverage any extra knowledge (e.g., relational knowledge) to enhance word representations.


\section{Solving Verbal Questions}\label{sec-framework}

In this section, we introduce our proposed framework to solve the verbal questions, which consists of the following three components.

\subsection{Classification of Question Types}\label{sec-algo-classifier}

The first component of the framework is a question classifier, which identifies different types of verbal questions. Since different types of questions usually have their unique ways of expressions, the classification task is relatively easy, and we therefore take a simple approach to fulfill the task. Specifically, we regard each verbal question as a short document and use the TF$\cdot$IDF features to build its representation. Then we train an SVM classifier with linear kernel on a portion of labeled question data, and apply it to other questions. The question labels include Analogy-I, Analogy-II, Classification, Synonym, and Antonym. We use the \emph{one-vs-rest} training strategy to obtain a linear SVM classifier for each question type.

\subsection{Embedding of Word-Senses and Relations}\label{sec-algo-embedding}

The second component of our framework leverages deep learning technologies to learn distributed representations for words (i.e. word embedding). Note that in the context of verbal question answering, we have some specific requirements on this learning process. Verbal questions in IQ tests usually consider the multiple senses of a word (and focus on the rare senses), and the complex relations among (polysemous) words, such as synonym and antonym relation. These challenges have exceeded the capability of standard word embedding technologies. To address this problem, we propose a novel approach that considers the multi-sense nature of words and integrate the relational knowledge among words (or their senses) into the learning process. In particular, our approach consists of two steps. The first step aims at labeling a word in the text corpus with its specific sense, and the second step employs both the labeled text corpus and the relational knowledge contained in dictionaries to simultaneously learn embeddings for both word-sense pairs and relations.

\subsubsection{Multi-Sense Identification}

First, we learn a single-sense word embedding by using the skip-gram method in \emph{word2vec}~\cite{mikolov2013distributed}.

Second, we gather the context windows of all occurrences of a word used in the skip-gram model, and represent each context by a weighted average of the pre-learned embedding vectors of the context words. We use TF$\cdot$IDF to define the weighting function, where we regard each context window of the word as a short document to calculate the document frequency. Specifically, for a word $w_{0}$, each of its context window can be denoted by $(w_{-N},\cdots,w_{0},\cdots,w_{N})$. Then we represent the window by calculating the weighted average of the pre-learned embedding vectors of the context words as below,
\begin{equation}
\xi=\frac{1}{2N}\sum_{i=-N, i\neq 0}^{N}{g_{w_i}v_{w_i}},
\end{equation}
where $g_{w_i}$ is the TF$\cdot$IDF score of $w_i$, and $v_{w_i}$ is the pre-learned embedding vector of $w_i$. After that, for each word, we use spherical $k$-means to cluster all its context representations, where cluster number $k$ is set as the number of senses of this word in the online dictionary.

Third, we match each cluster to the corresponding sense in the dictionary. On one hand, we represent each cluster by the average embedding vector of all those context windows included in the cluster. For example, suppose word $w_{0}$ has $k$ senses and thus it has $k$ clusters of context windows, we denote the average embedding vectors for these clusters as $\bar{\xi}_1,\cdots,\bar{\xi}_k$. On the other hand, since the online dictionary uses some descriptions and example sentences to interpret each word sense, we can represent each word sense by the average embedding of those words including its description words and the words in the corresponding example sentences. Here, we assume the representation vectors (based on the online dictionary) for the $k$ senses of $w_{0}$ are $\zeta_1,\cdots,\zeta_k$. After that, we consecutively match each cluster to its closest word sense in terms of the distance computed in the word embedding space, i.e.,
\begin{equation}\label{equ-match}
(\bar{\xi}_{i'}, \zeta_{j'}) = \mathop{\argmin}_{i,j=1,\cdots,k}{d(\bar{\xi}_i, \zeta_j)},
\end{equation}
where $d(\cdot,\cdot)$ calculates the Euclidean distance and $(\bar{\xi}_{i'}, \zeta_{j'})$ is the first matched pair of window cluster and word sense. Here, we simply take a greedy strategy. That is, we remove $\bar{\xi}_{i'}$ and $\zeta_{j'}$ from the cluster vector set and the sense vector set, and recursively run (\ref{equ-match}) to find the next matched pair till all the pairs are found. Finally, each word occurrence in the corpus is relabeled by its associated word sense, which will be used to learn the embeddings for word-sense pairs in the next step.

\subsubsection{Co-Learning Word-Sense Pair Representations and Relation Representations}\label{sec-algo-embedding-step2}

After relabeling the text corpus, different occurrences of a polysemous word may correspond to its different senses, or more accurately word-sense pairs.  We then learn the embeddings for word-sense pairs and relations (obtained from dictionaries, such as synonym and antonym) simultaneously, by integrating relational knowledge into the objective function of the word embedding learning model like skip-gram. We propose to use a function ${E}_{r}$ as described below to capture the relational knowledge.

Specifically, the existing relational knowledge extracted from dictionaries, such as synonym, antonym, etc., can be naturally represented in the form of a triplet (\emph{head}, \emph{relation}, \emph{tail}) (denoted by $(h_i,r,t_j) \in S$, where $S$ is the set of relational knowledge), which consists of two word-sense pairs (i.e. word $h$ with its $i$-th sense and word $t$ with its $j$-th sense), $h, t \in W$ ($W$ is the set of words) and a relationship $r \in R$ ($R$ is the set of relationships). To learn the relation representations, we make an assumption that relationships between words can be interpreted as translation operations and they can be represented by vectors. The principle in this model is that if the relationship $(h_i,r,t_j)$ exists, the representation of the word-sense pair $t_j$ should be close to that of $h_i$ plus the representation vector of the relationship $r$, i.e. $h_i+r$; otherwise, $h_i+r$ should be far away from $t_j$. Note that this model learns word-sense pair representations and relation representations in a unified continuous embedding space.

According to the above principle, we define ${E}_{r}$ as a margin-based regularization function over the set of relational knowledge $S$,

{\scriptsize
\begin{equation}
\label{eqn_Er}
{E}_{r}= \sum_{(h_i,r,t_j) \in  S } \sum_{(h^{'}, r, t^{'})\in S^{'}_{(h_i,r,t_j)} }\left [ \gamma +d(h_i+r,t_j) - d(h^{'}+r,t^{'}) \right ]_{+}.\nonumber
\end{equation}
}

Here $[X]_{+}$ denotes the positive part of $X$, $\gamma > 0$ is a margin hyperparameter, and $d(\cdot,\cdot)$ is the distance measure for the words in the embedding space. For simplicity, we again define $d(\cdot,\cdot)$ as the Euclidean distance. The set of corrupted triplets $S^{'}_{(h,r,t)}$ is defined as:
\begin{equation}
S^{'}_{(h_i,r,t_j)} = \left \{  (h ^{'}, r, t) \right \} \bigcup \left \{  (h , r, t^{'})  \right \},
\end{equation}
which is constructed from $S$ by replacing either the head word-sense pair or the tail word-sense pair by another randomly selected word with its randomly selected sense.

Note that the optimization process might trivially minimize ${E}_{r}$ by simply increasing the norms of word-sense pair representations and relation representations. To avoid this problem, we use an additional constraint on the norms, which is a commonly-used trick in the literature~\cite{bordes2011learning}. However, instead of enforcing the ${L}_{2}$-norm of the representations to 1 as used in~\cite{bordes2011learning}, we adopt a soft norm constraint on the relation representations as below:
\begin{equation}
{r}_{i} = 2  \sigma({x}_{i}) - 1,
\end{equation}
where $\sigma(\cdot)$ is the sigmoid function $\sigma(x_i) = 1/(1+e^{-x_i})$, ${r}_{i}$ is the $i$-th dimension of relation vector $r$, and ${x}_{i}$ is a latent variable, which guarantees that every dimension of the relation representation vector is within the range $(-1, 1)$.

By combining the skip-gram objective function and the regularization function derived from relational knowledge, we get the following combined objective ${J}_{r}$ that incorporates relational knowledge into the word-sense pair embedding calculation process,
\begin{equation}
\label{loss_r}
{J}_{r} = \alpha {E}_{r} - L,
\end{equation}
where $\alpha$ is the combination coefficient. Our goal is to minimize the combined objective ${J}_{r}$, which can be optimized using back propagation neural networks. 
By using this model, we can obtain the distributed representations for both word-sense pairs and relations simultaneously.


\subsection{Solvers for Each Type of Questions}\label{sec-algo-solvers}

\subsubsection{Analogy-I}

For the Analogy-I questions like ``$A$ is to $B$ as $C$ is to ?'', we answer them by optimizing:
\begin{equation}
D = \mathop{\argmax}_{i_b, i_a, i_c, i_{d'};D' \in T}{cos(v_{(B,i_b)} - v_{(A,i_a)} + v_{(C,i_c)}, v_{(D',i_{d'})})},
\end{equation}
where $T$ contains all the candidate answers, $cos$ means cosine similarity, and $i_b, i_a, i_c, i_{d'}$ are the indexes for the word senses of $B, A, C, D'$ respectively. Finally $D$ is selected as the answer.

\subsubsection{Analogy-II}

As the form of the Analogy-II questions is like ``$A$ is to ? as $C$ is to ?'' with two lists of candidate answers, we can apply an optimization method as below to select the best $(B, D)$ pair,
\begin{equation}
\mathop{\argmax}_{i_{b'}, i_a, i_c, i_{d'};B' \in T_1, D' \in T_2}{cos(v_{(B',i_{b'})} - v_{(A,i_a)} + v_{(C,i_c)}, v_{(D',i_{d'})})},
\end{equation}
where $T_1$, $T_2$ are two lists of candidate words. Thus we get the answers $B$ and $D$ that can form an analogical relation between word pair ($A$, $B$) and word pair ($C$, $D$) under a certain specific word sense combination.

\subsubsection{Classification}

For the Classification questions, we leverage the property that words with similar co-occurrence information are distributed close to each other in the embedding space. As there is one word in the list that does not belong to others, it does not have similar co-occurrence information with other words in the training corpus, and thus this word should be far away from other words in the word embedding space.

According to the above discussion, we first calculate a group of mean vectors $m_{i_{w_1},\cdots,i_{w_N}}$ of all the candidate words with any possible word senses as below,
\begin{equation}
m_{i_{w_1},\cdots,i_{w_N}} = \frac{1}{N}\sum_{w_j \in T}{v_{(w_j,i_{w_j})}},
\end{equation}
where $T$ is the set of candidate words, $N$ is the capacity of $T$, $w_j$ is a word in $T$; $i_{w_j} (j=1,\cdots,N; i_{w_j}=1,\cdots,k_{w_j})$ is the index for the word senses of $w_j$, and $k_{w_j} (j=1,\cdots,N)$ is the number of word senses of $w_j$. Therefore, the number of the mean vectors is $M=\prod_{j=1}^N k_{w_j}$. As both $N$ and $k_{w_j}$ are very small, the computation cost is acceptable. Then, we choose the word with such a sense that its closest sense to the corresponding mean vector is the largest among the candidate words as the answer, i.e.,
\begin{equation}
w = \mathop{\argmax}_{w_j \in T}\min_{i_{w_j};l=1,\cdots,M}{d(v_{(w_j,i_{w_j})}, m_l)}.
\end{equation}

\subsubsection{Synonym}

For the Synonym questions, we empirically explored two solvers. For the first solver, we also leverage the property that words with similar co-occurrence information are located closely in the word embedding space. Therefore, given the question word $w_q$ and the candidate words $w_i$, we can find the answer by the following optimization problem.
\begin{equation}
w = \mathop{\argmin}_{i_{w_q},i_{w_j};w_j \in T}{d(v_{(w_j,i_{w_j})}, v_{(w_q,i_{w_q})})},
\end{equation}
where $T$ is the set of candidate words. The second solver is based on the minimization objective of the translation distance between entities in the relational knowledge model (\ref{eqn_Er}). Specifically, we calculate the offset vector between the embedding of question word $w_q$ and each word $w_j$ in the candidate list. Then, we set the answer $w$ as the candidate word with which the offset is the closest to the representation vector of the synonym relation $r_s$, i.e.,
\begin{equation}
w = \mathop{\argmin}_{i_{w_q},i_{w_j};w_j \in T}\big| |v_{(w_j,i_{w_j})} - v_{(w_q,i_{w_q})}| - r_s\big|.
\end{equation}
In practice, we found the second solver performs better (the results are listed in Section \ref{sec-experiments}).

\subsubsection{Antonym}
Similar to solving the Synonym questions, we explored two solvers for Antonym questions as well. That is, the first solver (\ref{eqn_solver_first}) is based on the small offset distance between semantically close words whereas the second solver (\ref{eqn_solver_second}) leverages the translation distance between two words' offset and the embedding vector of the antonym relation. One might doubt on the reasonableness of the first solver given that we aim to find an answer word with opposite meaning for the target word (i.e. antonym). We explain it here that since antonym and its original word have similar co-occurrence information, based on which the embedding vectors are derived, thus the embedding vectors of both words with antonym relation will still lie closely in the embedding space.
\begin{equation}
\label{eqn_solver_first}
w = \mathop{\argmin}_{i_{w_q},i_{w_j};w_j \in T}{d(v_{(w_j,i_{w_j})}, v_{(w_q,i_{w_q})})},
\end{equation}
\begin{equation}
\label{eqn_solver_second}
w = \mathop{\argmin}_{i_{w_q},i_{w_j};w_j \in T}\big| |v_{(w_j,i_{w_j})} - v_{(w_q,i_{w_q})}| - r_a\big|,
\end{equation}
where $T$ is the set of candidate words and $r_a$ is the representation vector of the antonym relation. Again we found that the second solver performs better. Similarly, for skip-gram, only the first solver is applied.

\section{Experiments}\label{sec-experiments}

We conduct experiments to examine whether our proposed framework can achieve satisfying results on verbal comprehension questions.

\subsection{Data Collection}

\subsubsection{Training Set for Word Embedding}

We trained word embeddings on a publicly available text corpus named \emph{wiki2014}\footnote{\url{http://en.wikipedia.org/wiki/Wikipedia:Database_download}}, which is a large text snapshot from Wikipedia. After being pre-processed by removing all the \emph{html} meta-data and replacing the digit numbers by English words, the final training corpus contains more than 3.4 billion word tokens, and the number of unique words, i.e. the vocabulary size, is about 2 million.

\subsubsection{IQ Test Set}\label{sec-IQtestset}

According to our study, there is no online dataset specifically released for verbal comprehension questions, although there are many online IQ tests for users to play with. In addition, most of the online tests only calculate the final IQ scores but do not provide the correct answers. Therefore, we only use the online questions to train the verbal question classifier described in Section \ref{sec-algo-classifier}. Specifically, we manually collected and labeled 30 verbal questions from the online IQ test Websites\footnote{\url{http://wechsleradultintelligencescale.com/}} for each of the five types (i.e. Analogy-I, Analogy-II, Classification, Synonym, and Antonym) and trained an \emph{one-vs-rest} SVM classifier for each type. The total accuracy on the training set itself is 95.0\%. The classifier was then applied in the test set below.

We collected a set of verbal comprehension questions associated with correct answers from the published IQ test books, such as~\cite{Carter2005,carter2007ultimate,dan1993original,ken2002times}, and we used this collection as the test set to evaluate the effectiveness of our new framework. In total, this test set contains 232 questions with the corresponding answers.
The statistics of each question type are listed in Table~\ref{table:stats}.

\begin{table}
  \centering\footnotesize
  \vspace{-5pt}
  \caption{Statistics of the verbal question test set.}\label{table:stats}
  \begin{tabular}{c|c}
  \hline
  \hline
  Type of Questions & Number of questions \\
  \hline
  Analogy-I & 50 \\
  Analogy-II & 29 \\
  Classification & 53 \\
  Synonym & 51 \\
  Antonym & 49 \\
  \hline
  \textbf{Total} & 232\\
  \hline
  \end{tabular}\vspace{-15pt}
\end{table}
	
%

\subsection{Compared Methods}\label{sec-baseline}

In the following experiments, we compare our new relation knowledge powered model to several baselines.

{\bf Random Guess Model (RG).} Random guess is the most straightforward way for an agent to solve questions. In our experiments, we used a random guess agent which would select an answer randomly regardless what the question was. To measure the performance of random guess, we ran each task for 5 times and calculated the average accuracy.

{\bf Human Performance (HP).} Since IQ tests are designed to evaluate human intelligence, it is quite natural to leverage human performance as a baseline. To collect human answers on the test questions, we delivered them to human beings through Amazon Mechanical Turk, a crowd-sourcing Internet marketplace that allows people to participate Human Intelligence Tasks. In our study, we published five Mechanical Turk jobs, one job corresponding to one specific question type. The jobs were delivered to 200 people. To control the quality of the collected results, we took several strategies: (i) we imposed high restrictions on the workers - we required all the workers to be native English speakers in North American and to be Mechanical Turk Masters (who have demonstrated high accuracy on previous Human Intelligence Tasks on the Mechanical Turk marketplace); (ii) we recruited a large number of workers in order to guarantee the statistical confidence in their performances; (iii) we tracked their age distribution and education background, which are very similar to those of the overall population in the U.S. While we can continue to improve the design, we believe the current results already make a lot of sense.

{\bf Latent Dirichlet Allocation Model (LDA).} This baseline model leveraged one of the most classical distributional word representations, i.e. Latent Dirichlet Allocation (LDA)~\cite{blei2003latent}. In particular, we trained word representations using LDA on \emph{wiki2014} with the topic number 1000.

{\bf Skip-Gram Model (SG).} In this baseline, we applied the word embedding trained by skip-gram~\cite{mikolov2013distributed} (denoted by \textbf{SG-1}). In particular, when using skip-gram to learn the embedding on \emph{wiki2014}, we set the window size as 5, the embedding dimension as 500, the negative sampling count as 3, and the epoch number as 3. In addition, we also employed a pre-trained word embedding by Google\footnote{\url{https://code.google.com/p/word2vec/}} with the dimension of 300 (denoted by \textbf{SG-2}).

{\bf Glove}. This baseline algorithm uses another powerful word embedding model Glove~\cite{pennington2014glove}. The configurations of running \textbf{Glove} are the same with those in running \textbf{SG-1}.

{\bf Multi-Sense Model (MS).} In this baseline, we applied the multi-sense word embedding models proposed in~\cite{EricHuang2012,Tian2014,neelakantan-EtAl2014Nonparametric} (denoted by \textbf{MS-1}, \textbf{MS-2} and \textbf{MS-3} respectively). For \textbf{MS-1}, we directly used the published multi-sense word embedding vectors by the authors\footnote{\url{http://ai.stanford.edu/~ehhuang/}}, in which they set 10 senses for the top $5\%$ most frequent words. For \textbf{MS-2} and \textbf{MS-3}, we get the embedding vectors by the released codes from the authors using the same configurations as \textbf{MS-1}.

{\bf Relation Knowledge Powered Model (RK).} This is our proposed method in Section~\ref{sec-framework}. In particular, when learning the embedding on \emph{wiki2014}, we set the window size as 5, the embedding dimension as 500, the negative sampling count as 3 (i.e. the number of random selected negative triples in $S'$), and the epoch number as 3. We adopted the online Longman Dictionary as the dictionary used in multi-sense clustering. We used a public relation knowledge set, WordRep~\cite{gao2014wordrep}, for relation training.

\subsection{Experimental Results}

\subsubsection{Accuracy of Question Classifier}

We applied the question classifier trained in Section \ref{sec-IQtestset} on the test set in Table \ref{table:stats}, and got the total accuracy 93.1\%. For RG and HP, the question classifier was not needed. For other methods, the wrongly classified questions were also sent to the corresponding wrong solver to find an answer. If the solver returned an empty result (which was usually caused by invalid input format, e.g., an Analogy-II question was wrongly input to the Classification solver), we would randomly select an answer.

\begin{table}
  \centering\tiny
  \caption{Accuracy of different methods among different human groups.}\label{table:accuracy}
  \vspace{0pt}
  \begin{tabular}{c|ccccc|c}
  \hline
  \hline
  & Analogy-I & Analogy-II & Classification & Synonym & Antonym & \textbf{Total} \\
  \hline
  {\bf RG} & {\bf 24.60} & {\bf 11.72} & {\bf 20.75} & {\bf 19.27} & {\bf 23.13} & {\bf 20.51}\\
  \hline
  {\bf LDA} & {\bf 28.00} & {\bf 13.79} & {\bf 39.62} & {\bf 27.45} & {\bf 30.61} & {\bf 29.31}\\
  \hline
  {\bf HP} & {\bf 45.87} & {\bf 34.37} & {\bf 47.23} & {\bf 50.38} & {\bf 53.30} & {\bf 46.23}\\  		
  \hline
  {\bf SG} & & & & & & \\
  \hdashline
  SG-1 & {\bf38.00} & {\bf24.14} & {\bf37.74} & {\bf45.10} & {\bf40.82} & {\bf38.36}\\
  SG-2 & {\bf38.00} & {\bf20.69} & {\bf39.62} & {\bf47.06} & {\bf44.90} & {\bf39.66}\\
  \hline
  {\bf Glove} & {\bf45.09} & {\bf24.14} & {\bf32.08} & {\bf47.06} & {\bf40.82} & {\bf 39.03}\\
  \hline
  {\bf MS} & & & & & & \\
  \hdashline
  MS-1 & {\bf36.36} & {\bf19.05} & {\bf41.30} & {\bf50.00} & {\bf36.59} & {\bf38.67}\\
  MS-2 & {\bf40.00} & {\bf20.69} & {\bf41.51} & {\bf49.02} & {\bf40.82} & {\bf40.09}\\
  MS-3 & {\bf 17.65} & {\bf 20.69} & {\bf 47.17} & {\bf 47.06} & {\bf 30.61} & {\bf 36.73}\\
  \hline
  {\bf RK} & {\bf48.00} & {\bf34.48} & {\bf52.83} & {\bf60.78} & {\bf51.02} & {\bf50.86}\\
  \hline
  \end{tabular}\vspace{-10pt}
\end{table}

%

\subsubsection{Overall Accuracy}

Table~\ref{table:accuracy} demonstrates the accuracy of answering verbal questions by using all the approaches mentioned in Section \ref{sec-baseline}. From this table, we have the following observations: (i) RK can achieve the best overall accuracy than all the other methods. In particular, RK can raise the overall accuracy by about $4.63\%$ over HP. (ii) RK is empirically superior than the skip-gram models SG-1/SG-2 and Glove. According to our understanding, the improvement of RK over SG-1/SG-2/Glove comes from two aspects: multi-sense and relational knowledge. Note that the performance difference between MS-1/MS-2/MS-3 and SG-1/SG-2/Glove is not significant, showing that simply changing single-sense word embedding to multi-sense word embedding does not bring too much benefit. One reason is that the rare word-senses do not have enough training data (contextual information) to produce high-quality word embedding. By further introducing the relational knowledge among word-senses, the training for rare word-senses will be linked to the training of their related word-senses. As a result, the embedding quality of the rare word-senses will be improved. (iii) RK is empirically superior than the two multi-sense algorithms MS-1, MS-2 and MS-3, demonstrating the effectiveness brought by adopting less model parameters and using online dictionary in building the multi-sense embedding model. 

These results are quite impressive, indicating the potential of using machine to comprehend human knowledge and even achieve the comparable level of human intelligence.

\subsubsection{Accuracy in Different Question Types}

Table~\ref{table:accuracy} reports the accuracy of answering various types of verbal questions by each comparing method. From the table, we can observe that the SG and MS models can achieve competitive accuracy on some certain question types (like Synonym) compared with HP. After incorporating knowledge into learning word embedding, our RK model can improve the accuracy over all question types. Moreover, the table shows that RK can result in a big improvement over HP on the question types of Synonym and Classification, while its accuracy on the other question types is not so good as these two types.

To sum up, the experimental results have demonstrated the effectiveness of the proposed RK model compared with several baseline methods. Although the test set is not large, the generalization of RK to other test sets should not be a concern due to the unsupervised nature of our model.

\section{Conclusions}\label{sec-conclusions}

We investigated how to automatically solve verbal comprehension questions in IQ Tests by using the word embedding techniques in deep learning. In particular, we proposed a three-step framework: (i) to recognize the specific type of a verbal comprehension question by a classifier, (ii) to leverage a novel deep learning model to co-learn the representations of both word-sense pairs and relations among words (or their senses), (iii) to design dedicated solvers, based on the obtained word-sense pair representations and relation representations, for addressing each type of questions. Experimental results have illustrated that this novel framework can achieve better performance than existing methods for solving verbal comprehension questions and even exceed the average performance of the Amazon Mechanical Turk workers involved in the experiments.


\bibliographystyle{named}
\bibliography{VerbalIQ}

\end{document}